\documentclass[letterpaper,twocolumn]{article}
\usepackage[margin=1in]{geometry}
\usepackage[numbers]{natbib}
\usepackage{graphicx}
\usepackage{hyperref}
\usepackage{placeins}
\usepackage{subcaption}
\usepackage{xspace}
\usepackage{amsfonts, amsmath, amsthm, amssymb}
\usepackage{algorithm}
\usepackage{algpseudocode}
\usepackage{booktabs}
\usepackage{xr}
\usepackage[capitalize,noabbrev]{cleveref}

\newtheorem{theorem}{Theorem}

\makeatletter
\DeclareRobustCommand\onedot{\futurelet\@let@token\@onedot}
\def\@onedot{\ifx\@let@token.\else.\null\fi\xspace}
 \def\Eg{\emph{E.g}\onedot}
\def\ie{\emph{i.e}\onedot, } \def\Ie{\emph{I.e}\onedot}

\makeatother

\newcommand{\method}{\textsc{GFT}\xspace}

\title{Gradient-Free Training of Quantized Neural Networks}
\date{}

\author{Noa Cohen\thanks{Denotes equal contribution, work was done while employed at Bosch.}\\
Technion -- Israel Institute of Technology\\
\and
Omkar Joglekar\footnotemark[1] \quad \\
Technical University of Munich
\and 
Michal Moshkovitz \quad Vladimir Tchuiev \quad Shir Kozlovsky \quad Dotan Di Castro\\
Bosch Centre for Artificial Intelligence
}

\begin{document}

\maketitle

\begin{abstract}
Training neural networks requires significant computational resources and energy. 
Methods like mixed-precision and quantization-aware training reduce bit usage, yet they still depend heavily on computationally expensive gradient-based optimization.
In this work, we propose a paradigm shift: eliminate gradients altogether. 
One might hope that, in a finite quantized space, finding optimal weights without gradients would be easier—but we theoretically prove that this problem is NP-hard even in simple settings where the continuous case is efficiently solvable.
To address this, we introduce a novel heuristic optimization framework that avoids full weight updates and significantly improves efficiency.
Empirically, our method achieves performance comparable to that of full-precision gradient-based training on standard datasets and architectures, while using up to $3\times$ less energy and requiring up to $5\times$ fewer parameter updates.

\end{abstract}
\section{Introduction} \label{sec:intro}

Deep Neural Networks are crucial components in many tasks such as image recognition \citep{deng2009imagenet}, speech recognition \citep{mehrish2023review}, 
chat-bots \citep{brown2020language},  % radford2019language
control systems \citep{fujimoto2021minimalist}, 
games \citep{berner2019dota, van2016deep}, and
algorithmic trading \citep{wang2021survey}. 
While neural networks are widely deployed, they demand substantial computational and energy resources for training \citep{tripp2024measuring}. This not only increases the cost of building and maintaining complex GPU cluster environments \citep{McKinsey} but also raises the carbon footprint associated with both training and inference \citep{lacoste2019quantifying}. 

Furthermore, their resource-intensive nature often makes it impractical to run these models on local edge devices \citep{lin2022device}, limiting opportunities for data privacy in sensitive applications. 
Several directions have been explored to make these models scalable, such as Model Distillation \cite{gou2021knowledge}, %Distributed Training,  
Mixed-Precision Training \cite{micikevicius2017mixed}, Post Training Quantization \cite{nagel2020up} and Quantization Aware Training \cite{Wang23arxiv}. 
Unfortunately, all of these techniques rely on gradients as the primary driver of the optimization, which is computationally expensive. Moreover, they require updating all model parameters at each optimization step.

In this paper, we argue that for networks with low-bit quantized weights, where only a few discrete weight updates are possible, full gradient-based updates are overly complex and unnecessary.
We therefore propose a new optimization technique called \emph{Gradient Free Training} (GFT), which uses stochastic search to train quantized neural networks—entirely without gradient computations. Instead of backpropagating gradients, GFT computes a low-precision contribution tensor that estimates the impact of each weight on the model’s output errors, guiding the optimization process.
We evaluate our method on both image classification and language modeling—across diverse datasets and architectures. Our approach enables quantized networks to achieve performance comparable to their full-precision counterparts, while reducing energy consumption by up to $3\times$ and the number of parameter updates by up to $5\times$ during training.
The algorithm is implemented as a torch.autograd function and optimizer class that enables the reuse of our optimizer in future research for quantized neural networks optimization. This also enables us to run hybrid networks with independent optimizers in the same experiment.

We propose the \method algorithm, but this naturally leads to a broader question: could there exist a simpler algorithm for the quantized setting? Surprisingly, the answer is no. We prove that finding optimal weights in a finite quantized space is NP-hard, even in simple cases where the full-precision continuous version can be solved efficiently.

In summary, our main contributions are as follows:
\begin{enumerate}

    \item We introduce an algorithm for training quantized deep neural networks without relying on high-precision gradients, drawing on ideas from discrete optimization. Unlike traditional methods, it avoids costly full-precision computations and updates only a small subset of model parameters at each step. 
    \item Experimentally, we demonstrate that our method enables efficient training of bounded-bit networks, with minimal performance degradation compared to full-precision models, while significantly reducing memory, compute, and energy consumption during training.
    \item We prove that finding an optimal linear separator when the weights are bounded is NP-hard, even for linearly separable data. This stands in contrast to the well-known fact that the unconstrained problem is solvable in polynomial time.
\end{enumerate}
\section{Related Work} \label{sec:related}

Various studies have explored techniques for model optimization through quantization and discretization of weights and/or activations. Notable works including \citet{han2016deep, iandola2016squeezenet, NIPS2015_ae0eb3ee, venkataramani2014axnn, Zhu2016LRADNNHA, PAN201788}, propose methods that accelerate computation by reducing network parameters and connections. However, despite these advancements, these approaches still rely on full precision calculations during training.

In contrast, other approaches replace traditional multiplications altogether. Research such as \citet{lin2016neural, courbariaux2016binaryconnect, li2022ternary, zhu2017trained, courbariaux2016binarized, rastegari2016xnornet, Deng18nn} substitute original computations with binary logic operations and accumulations. For instance, Binary Weight Networks (BWN) and Ternary Weight Networks (TWN) limit the network weights to binary ($\{-1, +1\}$) and ternary ($\{-1, 0, +1\}$) spaces, respectively, thus eliminating multiplication operations entirely. These methods focus on discretizing weight matrices, significantly reducing computational requirements.

Further advancements, as those in \citet{rastegari2016xnornet}, extend these methods by restricting both weights and activation functions to binary values ($\{-1, +1\}$), allowing multiply-accumulate operations to be replaced with binary logic operations such as XNOR. Enhancements such as Gated XNOR Networks \citep{Deng18nn} introduce more sophisticated discretization frameworks, achieving performance comparable to full-precision methods with significantly lower memory and computational costs. 
Recent innovations include BitNet, introduced by \citep{Wang23arxiv}, a transformer-based architecture utilizing binary weights to reduce energy consumption, and a modification by \citep{Ma24arxiv}, which employs ternary weights while achieving performance close to that of 16-bit transformers.

Although these methods are pivotal in advancing low-compute, energy-efficient neural networks, they still rely on gradient-based optimization methods, often involving rounding or quantization steps. Our work diverges from this by presenting a novel training method that eliminates the dependency on gradient-based approaches altogether. This offers an end-to-end solution for low-compute, memory-efficient, and energy-efficient training. Additionally, our method can complement the aforementioned techniques to further improve the performance and efficiency of low-compute neural networks.

Another line of work explores zero-order methods \citep{liu2019signsgd,chen2023deepzero, malladi2023fine}, which eliminate the need for backpropagation by estimating gradients through zero-order techniques, or train without any gradient computation \citep{hinton2022forward,tripathi2020rso}, all in the context of full-precision models.
More recently, \citet{bar2025zoqo} proposed a method which adapts a zero-order method for quantized training.
While such approaches update all parameters in the network, our method prioritizes which weights to update, significantly reducing the computational cost per iteration.
\section{Preliminaries} \label{sec:preliminaries}

In this section we will describe the problem we focus on and the notation that will be used throughout the paper. We focus on the classical problem of feature-based supervised machine learning. Specifically, we consider a neural network with $L$ layers and a labeled dataset comprising $N$ data points, denoted as $D = \{x_1, \ldots, x_N\}$, with labels $y_1, \ldots, y_N \in \mathbb{R}^{d_L}$. Each sample is a vector containing $d_0$ features, expressed as $x_n = [x_{n,0}, \ldots, x_{n,d_0-1}] \in \mathbb{R}^{d_0}$, where $n = 1, \ldots, N$. A batch $b$ of size $B$ sampled from the dataset $D$ is indicated by $X_b$. The main objective in this setup is to learn the parameter matrices $W^1, \ldots, W^L$, where $W^l \in \mathbb{R}^{d_{l-1} \times d_l}$ for $l=1, \ldots, L$, that minimize a loss function $\mathcal{L} = \frac{1}{N}\sum_{n=1}^N \ell(y_n, x_n^{(L)})$, where $\ell(\cdot, \cdot): \mathbb{R}^{d_L} \times \mathbb{R}^{d_L} \rightarrow \mathbb{R}^+$.

The forward pass of the Neural Network is written as:
\begin{equation}
    \begin{split}
    v_b^{(l)} = X_b^{(l-1)} \cdot W^{(l)}
    %\quad \textrm{(linear layer forward pass $l$-th layer)} 
    \\
    X_b^{(l)} = \sigma^{(l)} \left(v_b^{(l)}\right)
    %\quad \textrm{(activation layer forward pass $l$-th layer)} 
    \end{split}
    \quad l=1,\ldots, L
    \label{eq:linear}.
\end{equation}
where $\sigma_l(\cdot): \mathbb{R}^{d_l} \rightarrow \mathbb{R}^{d_l}$ for $l=1, \ldots, L$ is the activation function of the $l$-th layer.

We follow the notation of \citet{haykin1998neural} for the backpropagation algorithm, which is defined using the \emph{backpropagated gradients }$\delta^{(l)} \in \mathbb{R}^{B \times d_l}$ for $l = 1, \ldots, L-1$.
Initially $\delta^{(L)}$ is set to equal the gradient of the loss at the output. 
Then the backpropogated gradients are defined recursively by
\begin{equation}
\delta^{(l-1)} = \sigma' (v^{(l-1)}) \odot \left[ \delta^{(l)} \cdot (W^{(l)})^T \right], 
     \quad 
 \label{eq:gradients}
\end{equation}
where $\odot$ is the standard Hadamard product.
For the rest of the paper, we will describe our algorithm adhering to this notation.

\section{Hardness of Learning Quantized NNs} \label{sec:theory}
In this section, we investigate the computational hardness of learning neural networks when the weights are restricted to come from a small, discrete set — for example, $\{-1, 1\}$, $\{-1, 0, 1\}$, or more generally, $\{-I, \dots, I\}$ for small integers $I$. 

At first glance, the learning problem might appear easier when the hypothesis class is finite. Indeed, when the number of possible weight configurations is finite, one could in principle enumerate all networks and select the optimal one. However, this approach is computationally infeasible in practice, and our focus is on efficient, polynomial-time, learning algorithms. 

Surprisingly, we show that efficient learning becomes harder under such constraints: even in very simple settings, the problem becomes computationally intractable. Specifically, we prove NP-hardness in the next theorem even when the data is linearly separable. This stands in sharp contrast to the real-valued case: when weights are allowed to take arbitrary real values, finding an optimal linear separator can be done efficiently \citep{shalev2014understanding}.

\begin{theorem} \label{th:npc}
Given a dataset \( \{(\mathbf{x}_i, y_i)\}_{i=1}^n \subset \mathbb{R}^d \times \{-1, 1\} \), the problem of deciding whether there exists a weight vector \( \mathbf{w} \in \{-1, 1\}^d \) such that
\[
y_i \cdot \mathbf{w}^\top \mathbf{x}_i > 0 \quad \forall i
\]
is NP-complete.
\end{theorem}
The proof is given in \cref{A:sec:NPH}.

\section{Method} \label{sec:method}
\begin{algorithm*}[htb!]
\caption{Top-K Gradient Free Neural Network Training  Algorithm}\label{alg:gradient_free}
\textbf{Input:} Labeled data: $D=\left(x_n,y_n\right)_{n=1}^{N} \in \mathbb{R}^{d_0}, \mathbb{R}^{d_L}$; Scheduler  $\text{SC}(t): \mathbb{N} \rightarrow \mathbb{N}$ which maps timestep to $k$ \\
\textbf{Parameters:} Number of training iterations $T$; Batch size $B$; Min flip probability $p_{min}$; Layer sizes $d_0, \ldots, d_L$; Max integer I \\
\textbf{Output:} Trained Weights ${W}^{(1)}, {W}^{(2)}, \ldots, {W}^{(L)}$

\begin{algorithmic}[1]
\State Randomly initialize $W^{(l)}\in \{-1,0, 1\}^{d_{l-1} \times d_l}$ \Comment{$\forall l=1, \ldots, L$}
\For{$t=1, \ldots, T$}
    \State $k \gets \text{SC}(t)$
    \State $\{C^{(1)}, \ldots, C^{(L)}\}, X_b^{(L)} \gets \text{ModifiedFwd}(D, B, \{{W}^{(1)}, {W}^{(2)}, \ldots, {W}^{(L)}\})$
\State $\delta^{(L)} \gets \text{Loss}\left( y_b, X_b^{(L)} \right)$ \Comment{Compute output $\delta$}
    \State // Backward pass
\For{$l=L, \ldots, 1$}
    \For{$b=1,\ldots,B$} \Comment{Compute tensor $\mathbf{B}^{(l)}$}
    \If{$\delta^{(l)}_{bo}c_{bio}^{(l)} < 0$} \Comment{An error exists}
    \State $\beta_{io}^{(l)} \gets \beta_{io}^{(l)}+sign(\delta_{bo}^{(l)} x^{(l-1)}_{bi}) $ \Comment{Summing up the contributions}
    \Else
    \State  $\beta_{io}^{(l)} \gets 0 $ 
    \EndIf
    \EndFor
    \State $\mathbf{B}^{(l)}_k\gets$ top-$k$ from $\mathbf{B}^{(l)}$
    \State $\mathbf{P}^{(l)}\gets \text{DynamicProbability}(\mathbf{B}^{(l)}_k, p_{min})$
    \For{$p_{io}^{(l)}$ in $\mathbf{P}^{(l)}$} 
        \State $r \sim \mathcal{U}[0, 1]$\Comment{Randomly flip the weight}
        \If{$r<p_{io}^{(l)}$}
        \State $w^{(l)}_{io} \gets w^{(l)}_{io}-sign(\beta^{(l)}_{io})$ 
        \State $w^{(l)}_{io} \gets Clip(w^{(l)}_{io}, \text{I}, -\text{I})$\Comment{Clip to valid range}
        \EndIf
    \EndFor
    \State $\delta^{(l-1)} \gets \delta^{(l)} \cdot {W^{(l)}}^\top$ \Comment{Preparing $\delta(\cdot)$ for the previous layer}
\EndFor
\EndFor
\end{algorithmic}
\end{algorithm*}

The essence of \method lies in replacing traditional gradient descent updates by leveraging the limited number of possible weight values in a quantized neural network. 
During the forward pass, a contribution matrix is computed, which captures the impact of each weight on the output. 
In the backward pass, the loss is propagated alongside this contribution matrix to determine which weights are most crucial to adjust. Instead of using gradients to update weights directly, \method updates the most important weights to the nearest quantized value that aligns with the desired direction of change, guided by their contributions to the loss. 
This method reduces the complexity of weight updates while effectively optimizing the model's performance in a quantized setting.
For simplicity, we start by explaining each of these stages for a single layer, and later in \cref{sec:deep}, we generalize this principle to a deep architecture.

Since the scope of this technique is focused on training quantized layers with few-bit weights, we restrict weight values to symmetric integer values such that the number of bits, $b$, determines the highest possible integer, $I$ by $I=2^{n-1}-1$.
\Eg setting $b=3$ results in the range of integers $\{-3,\ldots,3\}$, and setting $b=2$ gives ternary weights.

\subsection{The Contribution Tensor}
Observe a linear layer as in \eqref{eq:linear}. 
We define the \emph{contribution tensor} $C^{(l)}\in \mathbb{R}^{B\times d_{l-1}\times d_l}$ as an intermediate tensor, which signifies the contribution of each (data point, input feature, output feature) towards the final estimation. 
Formally, for each data point index $b\in \{1,\ldots,B\}$ in the current batch, input index $i\in \{1,\ldots,d_{l-1}\}$, and output index $o\in\{1,\ldots,d_{l}\}$, the element $c^{(l)}_{bio}$ of the matrix $C^{(l)}$ is defined as 
\begin{equation}
\label{eq:contibution_mat}
    c^{(l)}_{bio} = x^{(l-1)}_{bi}.w^{(l)}_{io}, 
\end{equation}
where, $x_{bi}^{(l-1)}$ is an element in the batch matrix $X_b^{(l-1)}$ and $w_{io}^{(l)}$ is an element in the weight matrix $W^{(l)}$. We note that the element $c^{(l)}_{bio}$ represents the contribution of the $i^{th}$ feature of the $b^{th}$ data point towards the $o^{th}$ output, in the $l^{th}$ layer of the network. Note that $w_{io}^{(l)}\in \{-I, \ldots, I\}$ and $x_{bi}^{(l-1)}$ may be in % $\{-1, +1\}$, $\{-1, 0, +1\}$, or 
$\mathbb{R}$. %, depending on the network precision. 
We note that if we sum the tensor $C^{(l)}$ along the input dimension, we recover the matrix $v^{(l)}$, as needed for a forward path. \Ie., for any data point $b$ and output $o$ 
\begin{equation} \label{eq:contributions_as_v}
    v^{(l)}_{bo} = \sum_{i=1}^{d_{l-1}} c^{(l)}_{bio} = \sum_{i=1}^{d_{l-1}} x^{(l-1)}_{bi}.w^{(l)}_{io}.
\end{equation}

Consider the case where the output of layer $l$ is negative, $v^{(l)}_{bo} < 0$, but the input of layer $l+1$, $x^{(l)}_{bi}$ is required to be \emph{positive} in order to be aligned with the label.
To correct the output, we need to modify some of the contributions where $c^{(l)}_{bio} < 0$, to make $v^{(l)}_{bo} > 0$. 
Similarly if, $v^{(l)}_{bo} > 0$ and we need $x^{(l)}_{bi} < 0$, then we need to change some contributions where $c^{(l)}_{bio} > 0$ to be negative.
Hence we use the contribution tensor in the backward pass to determine which weights might need to change.

\subsection{Back Propagation-like Update Rule}
Determining the optimal weight change that is consistent with all samples in a dataset or even a single batch is not always possible, as discussed in \cref{sec:theory}.
Furthermore, the combinatorial problem of identifying which weights to change can scale exponentially with the data dimension, hidden layer dimension, number of layers and the batch size.
To address this issue, we define a tensor $\mathbf{B}^{(l)}\in \mathbb{Z}^{d_{l-1}\times d_{l}}$, where $\mathbb{Z}$ is the set of integers. Each element of this tensor $\beta^{(l)}_{io}$, aggregates the contributions $\{c^{(l)}_{bio}\}_{b=1}^B$ for each weight $w^{(l)}_{io}$, based on the number of samples in $X^{(l-1)}_b$ that are negatively affected by this weight. If there is a mismatch between the sign of the output (or the backpropagated $\delta^{(l)}$) and the sign of $\{c^{(l)}_{bio}\}_{b=1}^B$, then an error exists. In this case $\beta^{(l)}_{io}$ is determined by:
\begin{equation}
    \beta^{(l)}_{io} = \sum_{b=1}^{B}sign(\delta_{bo}^{(l)} x^{(l-1)}_{bi}).
\end{equation}
Intuitively, the sign of $\beta^{(l)}_{io}$ denotes which direction $w^{(l)}_{io}$ needs to be updated and the magnitude signifies how many elements of the batch need this weight changed. Refer to lines 8-14 in \cref{alg:gradient_free}. 

Once $\mathbf{B}^{(l)}$ is created, we choose the top $k$ most destructive $\beta^{(l)}_{io}$, and mark the corresponding weights $w^{(l)}_{io}$ as candidates to be adjusted.
To calculate the probability of adjusting each of the weights, denoted by $p^{(l)}_{io}$, we normalize the $\beta^{(l)}$ values of all candidate weights to be in $[0,1]$, and change all values smaller than a minimal probability value $p_{\textrm{min}}$ to $p_{\textrm{min}}$.
\Ie., if we define with $K^{(l)}=\{(i,o) \hspace{0.5em}|\hspace{0.5em} \textrm{$|\beta_{io}^{(l)}|$ belongs to $k$ largest values} \}$, then, $\forall (i,o) \in d_{l-1}\times d_l$:
\begin{equation}
    w^{(l)}_{io} =
    \begin{cases}
    w^{(l)}_{io}-sign(\beta^{(l)}_{io}) & (i,o) \in K^{(l)} \textrm{ and w.p. $p^{(l)}_{io}$}\\
    w^{(l)}_{io}                        & \textrm{else}
    \end{cases}.
\end{equation}

Intuitively, the learning rule we propose identifies weights which effect outputs with problematic values. Hence, we should \emph{probably} change the weights so the output will more likely be consistent with the label. 
We note that practically, this problem is infeasible and equivalent to a random search of weights, such that we identify the weights that are most influential in terms of maximizing the correct output and flip a coin on whether to change them.

\subsection{Generalization to Deep Networks}
\label{sec:deep}

Inspired by the $\delta$ back-propagation update rule from \cite{haykin1998neural}, we extend our update rule to multiple layers using a similar technique of using $\delta^{(l)}$ as a surrogate signal for the inner gradient errors. Specifically, we define similarly $\delta^{(L)}=\mathcal{L}$ and $\delta^{(l-1)}$ as follows:
\begin{equation}
    \delta^{(l-1)} = \delta^{(l)} \cdot (W^{(l)})^T.
\end{equation}
We wish to highlight that if $\delta^{(l)}$ is an integer $\forall l=L, \ldots, l$, then $\delta^{(l-1)}$ will be an integer as well, since $W^{(l)}$ is ternary.

Furthermore, defining our algorithm this way, allows us to train \emph{hybrid networks}, where selective layers are ternary. For example, in a three layer Dense Neural Network, if layers $1$ and $3$ are full-precision (FP32) and layer $2$ is ternary, then the precision of $\delta^{(l)}$ for $l=1,2,3$ will be FP32, allowing us to optimize layers $1,3$ using standard optimizers such as AdamW, while optimizing layer $2$ using our technique. 
Note that our optimization still uses integers due to the discretization step in line 10 of~\cref{alg:gradient_free}.
The full algorithm is summarized in~\cref{alg:gradient_free}. 
The \texttt{ModifiedFwd} sub-routine can be found in \cref{A:fwd}. This sub-routine implements a standard forward pass along with the computation of the contribution tensors according to \cref{eq:contibution_mat}. 
The \texttt{DynamicProbability} sub-routine computes the probability of flipping for a particular element in a parameter proportional to it's amplitude in $\mathbf{B}^{(l)}$. The implementation of this sub-routine is also detailed in \cref{A:fwd}.

\begin{figure*}[t!]
    \centering
    \begin{subfigure}[t]{0.48\textwidth}
        \centering
        \includegraphics[width=\linewidth]{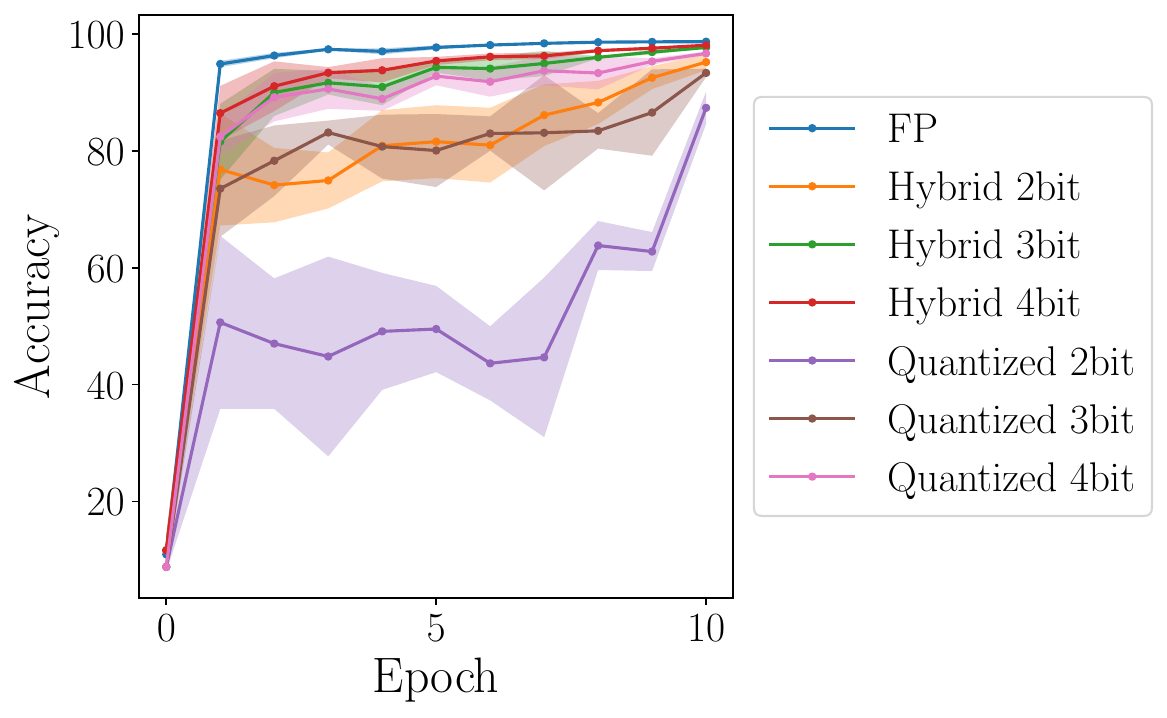}
        \caption{\textbf{DNN}}
        \label{fig:mnist_dnn}
    \end{subfigure}
    \hfill
    \begin{subfigure}[t]{0.48\textwidth}
        \centering
        \includegraphics[width=\linewidth]{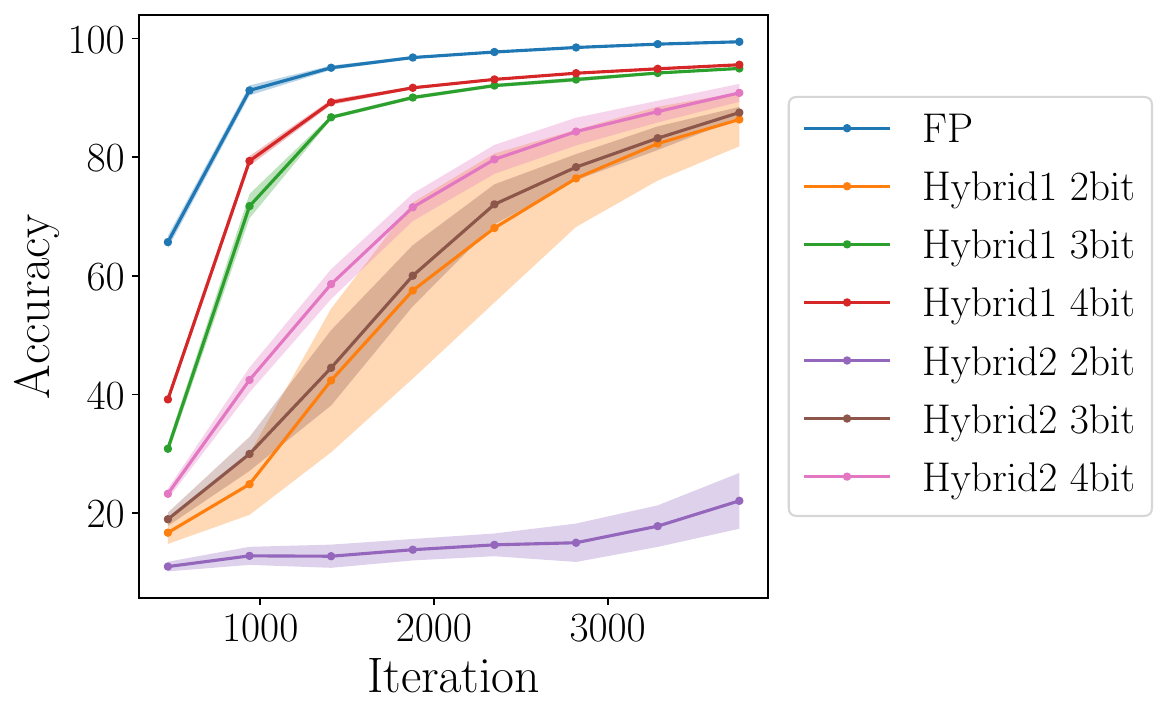}
        \caption{\textbf{ViT}}
        \label{fig:mnist_vit}
    \end{subfigure}
    \caption{\textbf{Comparison of model performance on MNIST validation set.} While the full-precision (FP) models achieve the highest accuracy, quantized models yield comparable performance, with accuracy decreasing in proportion to the degree of quantization. The 2-bit models exhibit largest standard deviation throughout the optimization process.}
    \label{fig:mnist_acc}
\end{figure*}

\subsection{Energy Consumption Estimation}
\label{sec:methods-energy}
A major advantage of our method is the reduced energy consumption, yet since there is no dedicated hardware to validate it on we refer to an estimation.
In previous methods such as BitNet \citep{Wang23arxiv} and PokeBNN \citep{zhang2022pokebnn}, the authors introduce a method to estimate the energy consumed by addition and multiplication operations on $45$nm and $7$nm technologies. Using their protocol, we attempt to estimate the amount of energy our optimization technique uses, as compared to a default implementation of AdamW in PyTorch, which we compare with.

First we estimate the energy use of a single AdamW optimization step. We refer to the algorithm described in the official PyTorch documentation \citep{pytorchAdamW}, assuming the default parameters. 
Assume that the whole model has $P_{FP32}$ parameters.  
A single step of AdamW consists of updating \emph{all} of the model parameters, along with applying weight decay and computing and normalizing the first and second moments. All of these tensors are of size $P$. In Lines 6-10 of the algorithm, we need to perform $8\times P_{FP32}$ FP32 multiplication and $2\times P_{FP32}$ FP32 addition operations. Assuming the Amsgrad option is disabled, we finally need to perform $2\times P_{FP32}$ FP32 multiplication and $P_{FP32}$ FP32 addition operations. In total, per step of AdamW, we perform $10\times P_{FP32}$ and $3\times P_{FP32}$ FP32 multiplication and addition operations respectively. Using the $7$nm values from \citet{Wang23arxiv}, that amounts to $14.62P_{FP32}$ [pJ] of energy per step.

Next, we estimate the energy consumed by our optimization step. Assume that we use a constant $k$ throughout the optimization, $P_{Q}$ is the number of ternary model parameters and that $p_{min}$ and $p_{max}$ are the minimum and maximum update probabilities respectively. For the sake of estimation, we assume that \emph{in expectation}, weights are updated with a $p_{change}=0.5*(p_{max}+p_{min})$. In our experiments $p_{min}=0.001$ and $p_{max}=1$ implying that $p_{change}\sim 0.5$. A single optimization step needs the computation of a topk, which has a complexity $O(P_{Q}+kP_{Q})$ FP32 addition operations, comparing our "flip probability tensor" element-wise with $p_{change}$, which is $O(P_{Q})$ FP32 addition operations and finally updating $p_{change}\times k \times P_{Q}$ weights, by adding or subtracting $1$ from each of them, for a total of $p_{change}\times k \times P_{Q}$ FP32 add operations. Assuming $k=0.75$ and $p_{change}=0.5$, which we use in all our experiments, we get a total energy use of $\sim 4.8P_{Q}$ [pJ], per update step. In the case of 3-bit and 4-bit quantization, we additionally need to perform $P_{Q}$ scalar multiplications with 3 and 7 respectively. Thus, the energy estimate for the 3 and 4 bit quantizations is $\sim 5.18P_{Q}$.

We use FP32 operation values to perform our estimation, but it is important to note that our optimization can be performed end to end using INT8. We do this to take into consideration the worst case and allow for optimization of hybrid networks.

\section{Experiments} \label{sec:experiments}
\begin{table*}[ht!]
    \caption{Results of training a 5-layer DNN on MNIST. Bold indicates best performance.}
    \centering
    \begin{tabular}{llllll}
    \toprule
       Model Name  & FP32 params [\#] & Accuracy [\%] & Updates [\#] & Bits [\#] & Energy [J] \\
       \midrule
       Full-Precision  & 53.6M & \textbf{98.7} & 105B & 1.72B & 1.54 \\
       Hybrid 4bit  & 3.22M & 98.1& 24.08B & 304.56M & 0.60 \\
       Hybrid 3bit & 3.22M & 97.7& 24.08B & 254.18M & 0.60 \\
       Hybrid 2bit & 3.22M & 95.2& 24.08B & 203.8M & 0.57 \\
       Quantized 4bit & 0 & 96.7& 18.91B & 214.4M & 0.54 \\
       Quantized 3bit & 0 & 93.4& 18.91B & 160.8M & 0.54 \\
       Quantized 2bit & 0 & 87.4& \textbf{18.91B} & \textbf{107.2M} & \textbf{0.50} \\
   \bottomrule
    \end{tabular}
\label{tab:mnist}
\end{table*}

\begin{table*}[ht]
    \caption{Results of training NanoGPT on Tiny Shakespeare. Bold indicates best performance.}
    \centering
    \begin{tabular}{llllll}
    \toprule
       Model Name  & FP32 params [\#] & Loss & Updates [\#] &  Bits [\#] & Energy [J] \\
       \midrule
       Full-Precision  & 10.8M & 1.57& 21.6B & 345.6M & 0.32 \\
       Hybrid-1 4bit  & 3.7M & \textbf{1.53}& 9.96B & 146.8M & 0.18 \\
       Hybrid-1 3bit & 3.7M & 1.61& 9.96B & 139.7M & 0.18 \\
       Hybrid-1 2bit & 3.7M & 1.93& 9.96B & 132.6M & 0.17 \\
       Hybrid-2 4bit  & 158.21K & 1.59& 4.15B & 47.62M & 0.11 \\
       Hybrid-2 3bit & 158.21K & 1.72& 4.15B & 36.98M & 0.11 \\
       Hybrid-2 2bit & 158.21K & 2.55& \textbf{4.15B} & \textbf{26.3M} & \textbf{0.10} \\
   \bottomrule
    \end{tabular}
\label{tab:ts}
\end{table*}

To assess the impact of \method as a training technique we examine its performance across different quantization levels. 
The primary focus is on understanding how the number of bits used for quantization affects the model's performance. We explore ternary, 3-bit, and 4-bit weights to evaluate the trade-offs between model precision and performance.
We apply each quantization scheme at different levels of quantization. Specifically, these levels differ in the number of layers where the FP32 parameters are replaced with their quantized versions. Each configuration is designed to test the behavior of the new training technique under varying conditions of quantization. Detailed descriptions of these configurations are provided in the following sections.
To ensure robustness and reliability, each configuration is evaluated over five independent runs, differing only in the random seed, and we report the mean performance across these runs (see \cref{fig:mnist_acc} and the Appendix for further statistics).

We evaluate our algorithm, \method, across three model architectures: Dense Neural Networks (DNNs), ViT-L~\cite{dosovitskiy2021imageworth16x16words} and nanoGPT~\cite{nanoGPT}; three datasets: MNIST~\cite{lecun1998mnist}, Imagenette~\cite{imagenette} (a 10-class subset of ImageNet), and Tiny Shakespeare~\cite {tinyshakespeare}; and two tasks: image classification and language modeling.
These models and datasets were selected to allow the evaluation of \method and how it trades the model accuracy, the energy consumption, and the update efficiency across different levels of quantization and across different tasks.

In all of our experiments, we linearly decay the value of $k$ in the \method optimizer from $0.75$ to $0$, with a decay rate proportional to $\frac{1}{T}$, where $T$ is the overall number of training iterations. In the training of the full-precision parameters, we use the default AdamW optimizer of PyTorch, with an initial learning rate of $0.0006$ and a weight decay set to $0.1$.
Quantized layers are initialized randomly with $90\%$ of weights $0$ and $10\%$ in \{-1,+1\}.
To enable easy usability with hybrid networks and seamless integration with the PyTorch framework, we implement a torch.autograd function and optimizer class used in all of the experiments. All experiments were conducted using an NVIDIA RTX 3090 GPU.

We report our findings across four metrics:
\begin{itemize}
    \item \textbf{Performance}: Top-1 validation accuracy for classification tasks, average validation loss for language modeling.
    \item \textbf{Number of bits}: As a proxy-metric to the memory required to store the model.
    \item \textbf{Number of updates}: Approximate number of variables updated throughout the training process.
    In gradient-based full-precision optimization, all model parameters are updated at every step, but in \method, only the worst $k\%$ parameters may be updated according to a probability that is proportional to the contribution matrix.
    \item \textbf{Energy Consumption}: Estimated energy consumption on $7$nm technology throughout the training process, as described in \cref{sec:methods-energy}. 
\end{itemize}

 % -------------------- DNNS --------------------
\subsection{Dense Neural Network Experiments}
We test the effectiveness of our optimization technique for DNNs, using the MNIST dataset. 
We apply three levels of quantization:
\begin{enumerate}
    \item Full-Precision: Standard FP32 model parameters trained using AdamW.
    \item Hybrid: Input and output layers of the model are FP32 while the hidden layers are quantized.
    \item Full-Quantized: All the layers of the model are quantized and optimized according to \method.
\end{enumerate}

The model consists of five fully-connected layers, with all hidden dimensions $4096$ such that the total number of trainable parameters is $53.6$M. An ablation study of these values is reported in the supplementary material.
All configurations are trained for $10$ epochs, with a batch size of $256$.

Looking at \cref{tab:mnist}, we observe that using our optimization technique, the Hybrid and Quantized 4-bit models attain accuracies of $98.1\%$ and $96.7\%$ respectively, that are comparable to the full-precision accuracy of $98.7\%$, while using only $0.6[J]$ and $0.54[J]$ of energy respectively. This marks a $\sim3\times$ reduction in energy consumption. We also update $\sim4.35\times$ less number of parameters.

 % -------------------- NanoGPT --------------------
\subsection{NanoGPT Experiments}
\label{lm-expts}
We train nanoGPT as described in \citet{nanoGPT}, on the Tiny Shakespeare dataset. Training is done using character tokenization, for $2000$ iterations, with a batch size of $128$ and a learning rate warmup for $10$ iterations.
We apply three levels of quantization:
\begin{enumerate}
    \item Full-Precision: All transformer parameters are FP32, trained using AdamW.
    \item Hybrid-1: Only the MLP in each attention block is quantized; the K, Q, V, and O projection matrices, the embedding lookup, positional encoding, and classification head are FP32. 
    \item Hybrid-2: The K, Q, V, O projection matrices and MLP in each attention block are quantized, but the embedding lookup, positional encoding, and classification head are FP32. 
\end{enumerate}

\begin{figure*}[ht!]
    \centering
    \makebox[0.3\textwidth][l]{%
        \fbox{\parbox[c][5.5cm][t]{0.3\textwidth}{
            \centering\textbf{FP}\par
            \raggedright{\fontsize{9}{10.5}\selectfont
            \footnotesize
            I shall be a constant to make the crown.\\
            DUKE VINCENTIO:\\
            You are a striken of transport, sir.
            MARIANA:\\
            I do beseech you, my lord.\\
            DUKE VINCENTIO:\\
            What say is the duke?\\
            LUCIO:\\
            Nay, I'll say the last of your grace of a mercy,
            that which he was born your profits to her entertain. \\
            LUCIO:\\
            Why, what is't well?\\
            ESCALUS:\\
        }}}
    }
    \hfill
    \makebox[0.3\textwidth][c]{%
        \fbox{\parbox[c][5.5cm][t]{0.3\textwidth}{
            \centering\textbf{Hybrid-1 4bit}\par
            \raggedright{\fontsize{9}{10.5}\selectfont
            \footnotesize
            The be the say the means to the truth him them. \\
            Provost: \\
            I have been the for the stay the had show shall spring,
            And when I did be hath stand my person,
            But not sweet us the so for the man, but were to my father's day
            The shall so soon of her her for you think. \\
            PAULINA: \\
            O, no, madam, and with with my profit words. \\
            POLIXENES: \\
            What say you will be cords are the the days. \\
        }}}
    }
    \hfill
    \makebox[0.3\textwidth][r]{%
        \fbox{\parbox[c][5.5cm][t]{0.3\textwidth}{
            \centering\textbf{Hybrid-1 2bit}\par
            \raggedright{\fontsize{9}{10.5}\selectfont
            \footnotesize
            Come he some sent the with hearst the him the prove of the wer for and\\
            And man thou see with his we seart sper word,\\
            Whe sham in all and with he would of the mines,\\
            My hear there ther the stand the and the can a from hear to hearthe poor the man othe do the my and bearted\\
            And with on a stall the this to the day and here to tween a come\\
            proce the the preath me the as abe the don of bert hey the world she come my were more to he have some,
        }}}
    }
    \caption{\textbf{Text generated by three of the NanoGPT models.} While all outputs resemble English-like text from the Shakespearean distribution, the 2-bit model deviates from the expected structure and appears less coherent.}
    \label{fig:boxes}
\end{figure*}
\begin{table*}[ht!]
    \centering
    \caption{Results of training a ViT-Lx16. Bold indicates best performance.}
    \begin{tabular}{llllll}
    \toprule
       Model Name  & FP32 params [\#] & Accuracy [\%] & Updates [\#] &  Bits [\#] & Energy [J] \\
       \midrule
        \multicolumn{6}{c}{MNIST} \\
       \midrule
       Full-Precision  & 37.87M & \textbf{99.5}& 151.48B & 1.2B & 2.21 \\
       Hybrid-1 4bit  & 12.67M & 95.6 & 68.82B & 506.24M & 1.26 \\
       Hybrid-1 3bit & 12.67M & 95.0 & 68.82B & 481.04M & 1.26 \\
       Hybrid-1 2bit & 12.67M & 86.4 & 68.82B & 455.84M & 1.22 \\
       Hybrid-2 4bit  & 70K & 90.1 & 27.5B & 153.44M & 0.78 \\
       Hybrid-2 3bit & 70K & 87.5 & 27.5B & 115.64M & 0.78 \\
       Hybrid-2 2bit & 70K & 22.1 & \textbf{27.5B} & \textbf{77.84M} & \textbf{0.72} \\
       \midrule
       \multicolumn{6}{c}{Imagenette} \\
       \midrule
       Full-Precision  & 38.33M & \textbf{90.8}& 191.65B & 1.23B & 2.8 \\
       Hybrid-1 4bit  & 13.23M &84.9 & 88.74B & 523.76M & 1.62 \\
       Hybrid-1 3bit & 13.23M &85.4 & 88.74B & 498.66M & 1.62 \\
       Hybrid-1 2bit & 13.23M &74.1 & 74.1B & 473.56M & 1.57 \\
       Hybrid-2 4bit  & 1.03M &77.0 & 38.72B & 182.16M & 1.04 \\
       Hybrid-2 3bit & 1.03M &70.4 & 38.72B & 144.86M & 1.04 \\
       Hybrid-2 2bit & 1.03M &35.4 & \textbf{38.72B} & \textbf{107.56M} & \textbf{0.97} \\
   \bottomrule
    \end{tabular}
\label{tab:vit}
\end{table*}

Evidently from \cref{tab:ts}, we observe that the 4-bit Hybrid configurations attain a final cross-entropy of $1.53$ and $1.59$ which are better and marginally worse than the full-precision loss of $1.57$. These results are also attained at about $1.78\times$ less energy consumption and updating $2.17\times$ less number of parameters.
Examples of text generated by three of the models are shown in~\Cref{fig:boxes}. While all outputs resemble English-like text from the Shakespearean distribution, the 2-bit model deviates from the expected structure and appears less coherent.

 % -------------------- VIT --------------------
\subsection{Vision Transformer Experiments}

We train a standard ViT-Lx16 model \cite{dosovitskiy2021imageworth16x16words} on the Imagenette dataset for $5000$ iterations, with a warm-up of $500$ iterations. On MNIST, we reduce the patch size to $4$ and the number of iterations and warmup to $4000$ and $100$ respectively.
All experiments were trained with a batch size of $256$. We apply three levels of quantization as in the NanoGPT experiments.

From \cref{tab:vit}, we observe that the full-precision model achieves higher accuracy on MNIST ($99.5\%$) than Hybrid-1-4bit ViT ($95.6\%$), as expected. 
However, this comes at the cost of approximately $5.5\times$ more parameter updates and $2.83\times$ higher energy consumption. 
Similar trends appear on Imagenette, where the accuracy gap is around $5.4\%$, and the full-precision model consumes roughly $3.7\times$ more energy and requires $4.9\times$ more parameter updates.

\section{Conclusion and Limitations} \label{sec:conclusion}

In this work we presented a gradient-free optimization rule for Quantized Neural Networks. We provided theoretical motivation for such a method, and empirically proved the advantages of our training protocol on a variety of datasets, architectures and tasks. Our method reduces energy consumption by about $3\times$, making the training of quantized networks more resource optimal, energy efficient and environment-friendly. We also implemented this optimization using default PyTorch's Autograd functionality, allowing users to use FP32 layers, interleaved with quantized ones while running optimization in a hybrid manner.
Currently, our method is limited due to the lack of dedicated hardware optimized for quantized and discretized numbers. We believe that continued progress in algorithms demonstrating high performance with low-bit representations will drive the development of dedicated hardware optimized for this paradigm.

\FloatBarrier
% \clearpage

\bibliographystyle{plainnat}
\bibliography{refs}

\clearpage
\appendix
\onecolumn

\section*{Appendices}
\section{Proof for Theorem 1} \label{A:sec:NPH}

We name the problem introduced in \cref{th:npc} \textsc{$\pm1$-Separable} and prove it is NP-complete.

\begin{proof}

To show \textsc{$\pm1$-Separable} is NP-complete we first show it is in NP, and then propose a polynomial-time reduction from \textsc{3-SAT} to \textsc{$\pm1$-Separable}.

Given a weight vector \(\mathbf{w}\), calculating \(y_i \cdot \mathbf{w}^\top \mathbf{x}_i > 0\) for each \(i \in \{1,...,n\}\) and comparing it with \(0\) can be performed in \(O(dn)\), proving that the problem can be verified in polynomial time, therefore it is in NP.

To prove it is NP-hard, we reduce from the NP-complete problem \textsc{3-SAT} which for completeness we describe next. Let \( \phi \) be a 3-CNF formula over Boolean variables \( z_1, \dots, z_m \), consisting of clauses \( C_1, \dots, C_n \), where each clause \( C_i = (l_{i1} \vee l_{i2} \vee l_{i3}) \), and each literal \( l_{it} \in \{z_k, \neg z_k\} \). In the problem of \textsc{3-SAT} one should decide whether there exists an assignment which satisfies all clauses in \( \phi \).

\paragraph{Reduction.}
For the reduction, given a 3-CNF formula \( \phi \), we first eliminate any clause that contains both a variable and its negation, since these are always satisfied and can be removed without changing the satisfiability of the formula. 
We denote by \(m\) the number of variables in \( \phi \), and set \(d=m+1\). 
For each remaining clause \( C_i \), we define a data point \( (\mathbf{x}_i, y_i) \in \mathbb{R}^d \times \{-1,1\} \) as follows:
\begin{itemize}
    \item \( y_i = 1 \)
    \item For each feature \( j = 1, \dots, d-1 \), set:
    \[
    x_{i,j} = 
    \begin{cases}
    1 & \text{if } z_j \in C_i \\
    -1 & \text{if } \neg z_j \in C_i \\
    0 & \text{otherwise}
    \end{cases}
    \]
    \item For the last feature, set 
    \[x_{i,d} = 2\]
\end{itemize}

We observe that the reduction is done in polynomial time since eliminating the redundant clauses can be done in linear time by checking for each clause whether a variable appears both positively and negatively, and that \( \{(\mathbf{x}_i, y_i)\}_{i=1}^n \) can be constructed in polynomial time.

\paragraph{Correctness.}
To show that the 3-CNF formula \( \phi \) is satisfiable if and only if there exists a weight vector \( \mathbf{w} \in \{-1, 1\}^d \) such that all constructed examples satisfy \( 1 \cdot \mathbf{w}^\top \mathbf{x}_i > 0 \) we prove each direction separately.

\begin{itemize}
    \item 
The 3-CNF formula \( \phi \) is satisfiable \( \Rightarrow \) The dataset is separable by some weight vector \( \mathbf{w} \in \{-1, 1\}^d \):

Let \(\alpha\) be the satisfying assignment such that \(\phi(\alpha)=\text{True}\). 
We interpret the first \(d-1\) features of the weight vector as an assignment for the \(d-1\) Boolean variables hence set them to be:
\[
w_j = 
\begin{cases}
1 & \text{if }  \alpha_j = \text{True} \\
-1 & \text{if }  \alpha_j = \text{False}
\end{cases}
 \quad \forall j \in {1,...,d-1}
\]
and choose \(w_d=1\).

Under this interpretation, the inner product \( \mathbf{w}^\top \mathbf{x}_i \) evaluates to a sum of contributions from the three literals in \( C_i \) and \(2\):
\[ 
    \mathbf{w}^\top \mathbf{x}_i = \sum_{j=1}^{d-1}w_jx_{ij} + 2 = w_ax_{ia}+w_bx_{ib}+w_cx_{ic}+2
\]
where the second equations holds since all values in the summation \(\sum_{j=1}^{d-1}w_jx_{ij}\) are \(0\) except for the \(3\) that correspond to the variable in the three literals denoted by indices \(\{a,b,c\}\).
Each literal contributes \( +1 \) if the literal is satisfied, as it is satisfied either if the literal is a positive literal and the assignment is True (\ie \(x=1\) and \(w=1\)), or if the literal is a negative literal and the assignment is False (\ie \(x=-1\) and \(w=-1\)).
In a similar manner, each literal contributes \( -1 \) if the literal is not satisfied, as it is not satisfied either if the literal is a positive literal and the assignment is False (\ie \(x=1\) and \(w=-1\)), or if the literal is a negative literal and the assignment is True (\ie \(x=-1\) and \(w=1\)). 

Thus, if the clause is satisfied (i.e., at least one literal is true), the sum of \(w_ax_{ia}+w_bx_{ib}+w_cx_{ic}\) is either \( -1 \), \( 1 \) or \( 3 \).
In all three cases, when adding \(2\) we get that the separability by the weight vector $\mathbf{w}$ holds as \( \mathbf{w}^\top \mathbf{x}_i \geq -1 + 2 > 0 \). 

\item
A satisfiable weight vector \( \mathbf{w} \in \{-1, 1\}^d \) exists \( \Rightarrow \) The 3-CNF formula \( \phi \) is satisfiable:

Let \(\mathbf{w}\) be the weight vector which satisfies the equation \(
y_i \cdot \mathbf{w}^\top \mathbf{x}_i > 0 \quad \forall i\).
We choose the assignment \(\alpha\) of the \(m=d-1\) variables to be:
\[
\alpha_k = 
\begin{cases}
\text{True} & \text{if } w_k = 1 \\
\text{False} & \text{if } w_k = -1
\end{cases}
 \quad \forall k \in {1,...,m}
\]

Assume by contradiction that there exists a clause \(C_i\) which is not satisfied by \(\alpha\), meaning that all three terms of the the literals in the clause are False hence contribute \(-1\) in the inner product \( \mathbf{w}^\top \mathbf{x}_i \).
Thus, 
\[
\mathbf{w}^\top \mathbf{x}_i = -3 + 2w_d
\]
In both possible cases of \(w_d = 1\) and \(w_d = -1\) we get a contradiction as \(-1 \leq 0\) and \(-5 \leq 0\).
Since our assumption leads to a contradiction, there is no clause which is not satisfied by \(\alpha\), and the 3-CNF formula \( \phi \) is satisfiable.
\end{itemize}

Hence, the problem is in NP and NP-hard, and is therefore NP-complete.
\end{proof}
\section{Additional Results} \label{A:sec:res}

\begin{figure}[ht]
    \centering
    \begin{subfigure}[t]{0.48\textwidth}
        \centering
        \includegraphics[width=\linewidth]{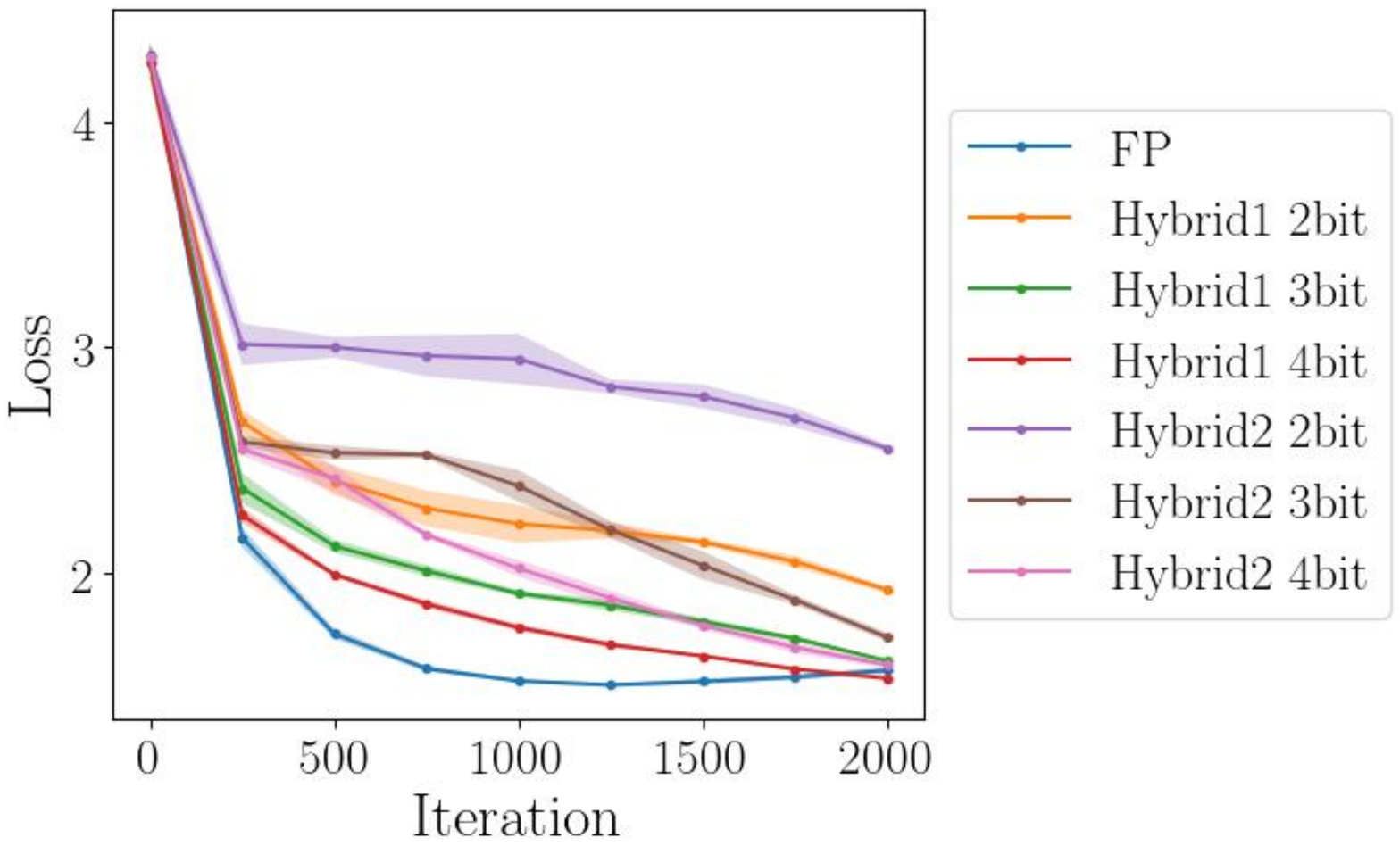}
        \caption{\textbf{NanoGPT trained on Tiny Shakespeare}}
        \label{fig:nanogpt}
    \end{subfigure}
    \hfill
    \begin{subfigure}[t]{0.48\textwidth}
        \centering
        \includegraphics[width=\linewidth]{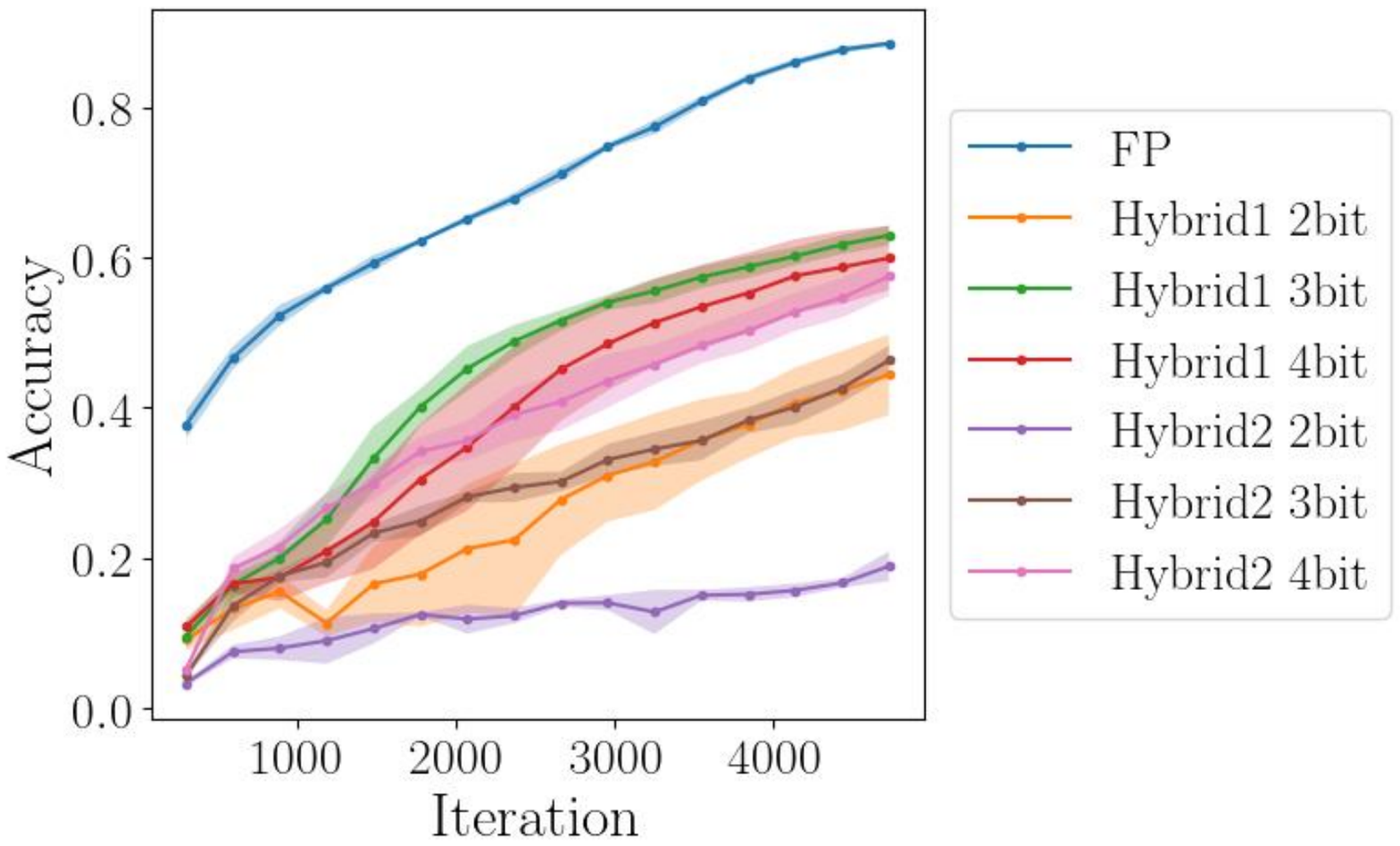}
        \caption{\textbf{ViT trained on Imagenette}}
        \label{fig:imagenette}
    \end{subfigure}
    \caption{\textbf{Comparison of model performance.} Full-precision (FP) models achieve best performance across both validation sets. While quantized models achieve comparable results on Tiny Shakespeare, their performance on Imagenette exhibits a more noticeable gap.}
    \label{fig:add_res}
\end{figure}

To provide a more comprehensive understanding of the variability in performance between different quantized models, the standard deviations corresponding to the mean values presented in the main text are detailed in \cref{tab:dnn-stats,tab:nanogpt-stats,tab:vit-stats}. 
The means and standard deviations were computed over five independent runs, revealing a trend in which more aggressive quantization leads to greater variance in overall performance.

\begin{table}[ht]
    \caption{Accuracy across DNN models trained on MNIST.}
    \centering
    \begin{tabular}{ll}
    \toprule
       Model Name  & Accuracy \\
       \midrule
       Full-Precision  & 98.7 $\pm$ 0.033 \\
       Hybrid 4bit     & 98.1 $\pm$ 0.033 \\
       Hybrid 3bit     & 97.7 $\pm$ 0.174 \\
       Hybrid 2bit     & 95.2 $\pm$ 1.364 \\
       Quantized 4bit  & 96.7 $\pm$ 0.315 \\
       Quantized 3bit  & 93.4 $\pm$ 0.719 \\
       Quantized 2bit  & 87.4 $\pm$ 2.780 \\
   \bottomrule
    \end{tabular}
\label{tab:dnn-stats}
\end{table}

\begin{table}[ht]
    \centering 
    \caption{Loss values across NanoGPT models trained on Tiny Shakespeare.}
    \begin{tabular}{ll} 
    \toprule
       Model Name & Loss \\
       \midrule
       Full-Precision & 1.57 $\pm$ 0.014 \\
       Hybrid-1 4bit  & 1.53 $\pm$ 0.004 \\
       Hybrid-1 3bit  & 1.61 $\pm$ 0.004 \\
       Hybrid-1 2bit  & 1.93 $\pm$ 0.015 \\
       Hybrid-2 4bit  & 1.59 $\pm$ 0.016 \\
       Hybrid-2 3bit  & 1.72 $\pm$ 0.019 \\
       Hybrid-2 2bit  & 2.55 $\pm$ 0.017 \\
   \bottomrule
    \end{tabular}
\label{tab:nanogpt-stats}
\end{table}

\begin{table}[ht]
    \centering
    \caption{Accuracy across ViT-Lx16 models.}
    \begin{tabular}{lll}
    \toprule
       Model Name & \multicolumn{2}{c}{Dataset} \\
    \cmidrule(lr){2-3}
        & MNIST & ImageNette \\
       \midrule

       Full-Precision & 99.5 $\pm$ 0.0005 & 90.8 $\pm$ 0.003 \\
       Hybrid-1 4bit  & 95.6 $\pm$ 0.002 & 95.6 $\pm$ 0.043 \\
       Hybrid-1 3bit  & 95.0 $\pm$ 0.002 & 85.4 $\pm$ 0.013 \\
       Hybrid-1 2bit  & 86.4 $\pm$ 0.045 & 74.1 $\pm$ 0.054 \\
       Hybrid-2 4bit  & 90.1 $\pm$ 0.015 & 77.0 $\pm$ 0.026 \\
       Hybrid-2 3bit  & 87.5 $\pm$ 0.009 & 70.4 $\pm$ 0.022 \\
       Hybrid-2 2bit  & 22.1 $\pm$ 0.047 & 35.4 $\pm$ 0.020 \\
   \bottomrule
    \end{tabular}
\label{tab:vit-stats}
\end{table}

\subsection{Network Depth Ablation on MNIST}
 
We assessed the effect of varying the number of layers on the performance of DNN models trained using \method when applying ternary (2Bit) weights, on the MNIST dataset. 
We tested configurations with 2, 3, 5, 7, and 10 layers, while the number of parameters in the hidden layer was fixed at 4096 for all configurations. 
The results are summarized in Table~\ref{tab:layers}, showing that the performance increases and decreases for all models with more than 5 layers.

\begin{table}[ht]
  \caption{Network Depth Ablation on MNIST}
  \label{tab:layers}
  \centering
  \begin{tabular}{llllll}
    \toprule
    & 2 Layers & 3 Layers & 5 Layers & 7 Layers & 10 Layers \\
    \midrule
    Accuracy & 84.46\% & 88.82\% & 89.24\% & 88.53\% & 88.47\% \\
    \bottomrule
  \end{tabular}
\end{table}
\FloatBarrier
\section{Subroutines of the \method algorithm}
\label{A:fwd}
This section details the implementation of the subroutines called by \cref{alg:gradient_free} in the main text. 
ModifiedFwd (\cref{alg:fwd}), is a modification of the known forward path which calculates the contribution tensor $C^{(l)}$ for each layer $l=1,\ldots, L$. 
DynamicProbability (\cref{alg:dyn}) transforms the matrix $\mathbf{B}^{(l)}\in \mathbb{Z}^{d_{l-1}\times d_{l}}$, representing the number of negatively affected samples per weight, into a matrix of probabilities for flipping each weight.

\begin{algorithm*}[!ht]
\caption{ModifiedFwd}\label{alg:fwd}
\begin{algorithmic}[1]
\Require \noindent Data with labels: $D=\left(x_n,y_n\right)_{n=1}^{N} \in \mathbb{R}^{d_0}, \mathbb{R}^{d_L}$; Batch size $B$;Model weights ${W}^{(1)}, {W}^{(2)}, \ldots, {W}^{(L)}$

\Ensure $C^{(l)} \forall l\in 1\ldots L$; $X_b^{(L)}$
\State Sample a batch $(X_b^{(0)}, y_b) \in D$ of size $B$.
\State // Forward pass
\For{$l=1, \ldots, L$} \Comment{Forward pass}
    \State Compute $C^{(l)}$ \Comment{Refer to Eq~\ref{eq:contibution_mat}}
    \State $v^{(l)} \gets \sum_{i=1}^{d_{l-1}} c^{(l)}_{bio}$ 
    \State $X_b^{(l)} \gets \sigma_f \left(v^{(l)} \right)$ 
\EndFor

\Return $\{C^{(1)}, \ldots, C^{(L)}\}$, $X_b^{(L)}$
\end{algorithmic}
\end{algorithm*}
\begin{algorithm*}[!ht]
\caption{DynamicProbability}\label{alg:dyn}
\begin{algorithmic}[1]
\Require \noindent top $k^{(l)}$ of $\mathbf{B}^{(l)}$, $\mathbf{B}^{(l)}(k^{(l)})$; minimum probability of flipping $p_{min}$

\Ensure Dynamic probability matrix $\mathbf{P}^{(l)}$

\If{$\mathbf{B}^{(l)}<0$ for all elements}

    \Return $zeros(size(\mathbf{B}^{(l)}))$
\EndIf

\State $b_{max}\gets max\{\mathbf{B}^{(l)}\}$
\State $b_{min}\gets min\{\mathbf{B}^{(l)}\}$
\State $\mathbf{P}^{(l)}\gets (\mathbf{B}^{(l)}-b_{min})/(b_{max}-b_{min})$

\Return $Clip(\mathbf{P}^{(l)}, 1, p_{min})$

\end{algorithmic}
\end{algorithm*}

\end{document}